\documentclass{article}
 
\usepackage[margin=1in]{geometry}
\usepackage[utf8]{inputenc}
\usepackage[T1]{fontenc}
\usepackage{lmodern}
\usepackage{amsmath,amsfonts,amssymb}
\usepackage{booktabs}
\usepackage{graphicx}
\usepackage{subcaption}
\usepackage{microtype}
\microtypesetup{expansion=false}
\usepackage{float}
\usepackage{tabularx}
\usepackage{enumitem}
\usepackage{comment}
\usepackage{natbib}
\usepackage{xcolor}
\usepackage{authblk}
\usepackage{tikz}
\usepackage{mathtools}
\usetikzlibrary{arrows.meta, positioning}
\usepackage[pdfpagemode=UseOutlines]{hyperref}
\hypersetup{hidelinks}
 
\definecolor{darkblue}{RGB}{25,55,100}
\definecolor{lightblue}{RGB}{225,238,250}
\definecolor{lightred}{RGB}{255,235,235}
\definecolor{lightgreen}{RGB}{230,245,235}
\definecolor{lightyellow}{RGB}{255,248,220}
\definecolor{darkgreen}{RGB}{35,120,75}
\definecolor{darkred}{RGB}{170,50,50}
\definecolor{darkorange}{RGB}{190,115,30}
\definecolor{draftgray}{HTML}{555555}
 
\newcolumntype{Y}{>{\raggedright\arraybackslash}X}
\newcommand{\code}[1]{\texttt{#1}}

\title{Automated Brick Schema Knowledge Graph Completion Across Buildings}

\author[1]{Daniel Ma}
\author[2]{Jalil Huseynov}
\author[2]{Xinyi Pan}
\author[2]{Shaked Shalev}
\author[2]{Zhang Chi}
\author[3]{Ai Yamamoto}
\author[3]{Sohei Arisaka}
\affil[1]{Northeastern University}
\affil[2]{National University of Singapore}
\affil[3]{Kajima Corporation}
 
\date{}
\begin{document}
\maketitle

\begin{abstract}
The Brick schema represents building equipment, sensors, spaces, and the relationships between them as typed, multi-relational graphs, providing machine-readable metadata that supports generalizable control applications. Current methods for producing complete Brick models require extensive manual examination of building schematics and point lists. We observe that as Brick models are knowledge graphs, inferring missing relations is naturally posed as a knowledge graph completion problem. To the best of our knowledge, this is the first application of knowledge graph completion to Brick model generation. We compare three methods, gradient-boosted trees, Pairwise Brick Relation Ranking, and CompGCN, on the relations \code{feeds}, \code{hasPart}, and \code{hasPoint}, which account for the majority of the schematic-reading work in building a Brick model by hand. Models are evaluated on held-out MORTAR buildings and out-of-distribution Kaggle and Building Time Series data. Although CompGCN is the strongest performer on held-out buildings under metrics resistant to class imbalance, leading macro-F1 and AUPRC, no method generalizes uniformly across datasets or relations.
\end{abstract}

\section{Introduction}
\label{sec:introduction}
Modern commercial buildings are equipped heavily. A single site may have thousands of devices and tens of thousands more data points, most exposed by a Building Management System (BMS) that controls heating, lighting and electrical equipment. This data is the foundation of smart building applications, including fault detection, demand response, and occupant comfort optimization \citep{pritoni2021metadata}. However, data alone is not enough. A time series named AHU3\_SAT\_SP doesn't tell software that AHU3 is an air handling unit, or that SAT\_SP is a supply air temperature setpoint or which zones it serves. Applications need metadata to know what the data means or how it relates to equipment and space.

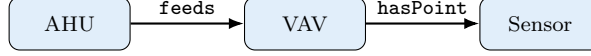
\begin{figure}[t]
\centering
\begin{tikzpicture}[
    box/.style={
        draw,
        rounded corners,
        fill=lightblue,
        minimum height=0.7cm,
        minimum width=1.6cm,
        align=center,
        font=\footnotesize
    },
    arrow/.style={-{Latex[length=2mm]}, thick}
]
\node[box] (ahu) {AHU};
\node[box, right=1.5cm of ahu] (vav) {VAV};
\node[box, right=1.5cm of vav] (sensor) {Sensor};

\draw[arrow] (ahu) -- node[above, font=\footnotesize]
    {\texttt{feeds}} (vav);

\draw[arrow] (vav) -- node[above, font=\footnotesize]
    {\texttt{hasPoint}} (sensor);
\end{tikzpicture}

\caption{A two-hop Brick graph fragment. Our completion targets are the typed,
directed relations \texttt{feeds}, \texttt{hasPart}, and \texttt{hasPoint}.}
\label{Fig:BrickPath}

\vspace{-6pt}
\end{figure}

Brick Schema is an open ontology that provides this metadata \citep{balaji2016brick}. A Brick model represents a building as a graph in which entities such as equipment, points, and spaces are instances of Brick classes, while relationships are typed directed edges such as \code{feeds}, \code{hasPart}, and \code{hasPoint}. This standard representation lets applications query entities by type and relationship rather than building-specific point names, enabling the same application to run across buildings with Brick models. Figure~\ref{Fig:BrickPath} shows a simple example.

\begin{figure}[t]
  \centering
  \edef\figh{\the\dimexpr0.23\linewidth\relax}%
  \begin{subfigure}[b]{0.37\linewidth}
    \centering
    \includegraphics[height=\figh]{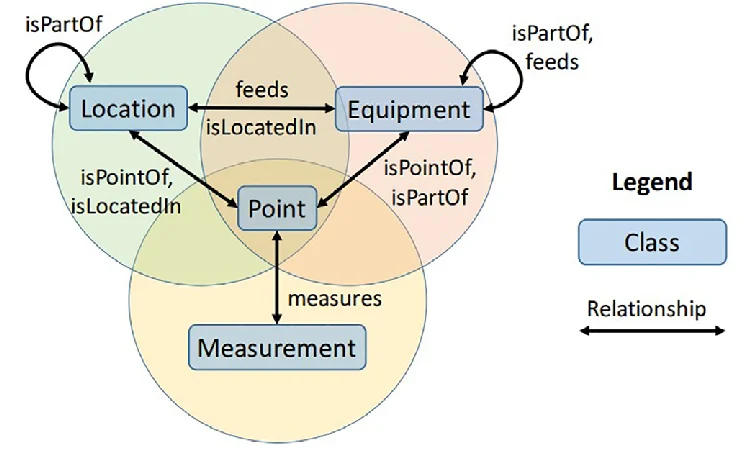}
    \caption{A Brick graph fragment.}
    \label{fig:brick}
  \end{subfigure}\hfill
  \begin{subfigure}[b]{0.56\linewidth}
    \centering
    \makebox[\linewidth]{%
      \raisebox{-0.5\height}{\includegraphics[height=\figh]{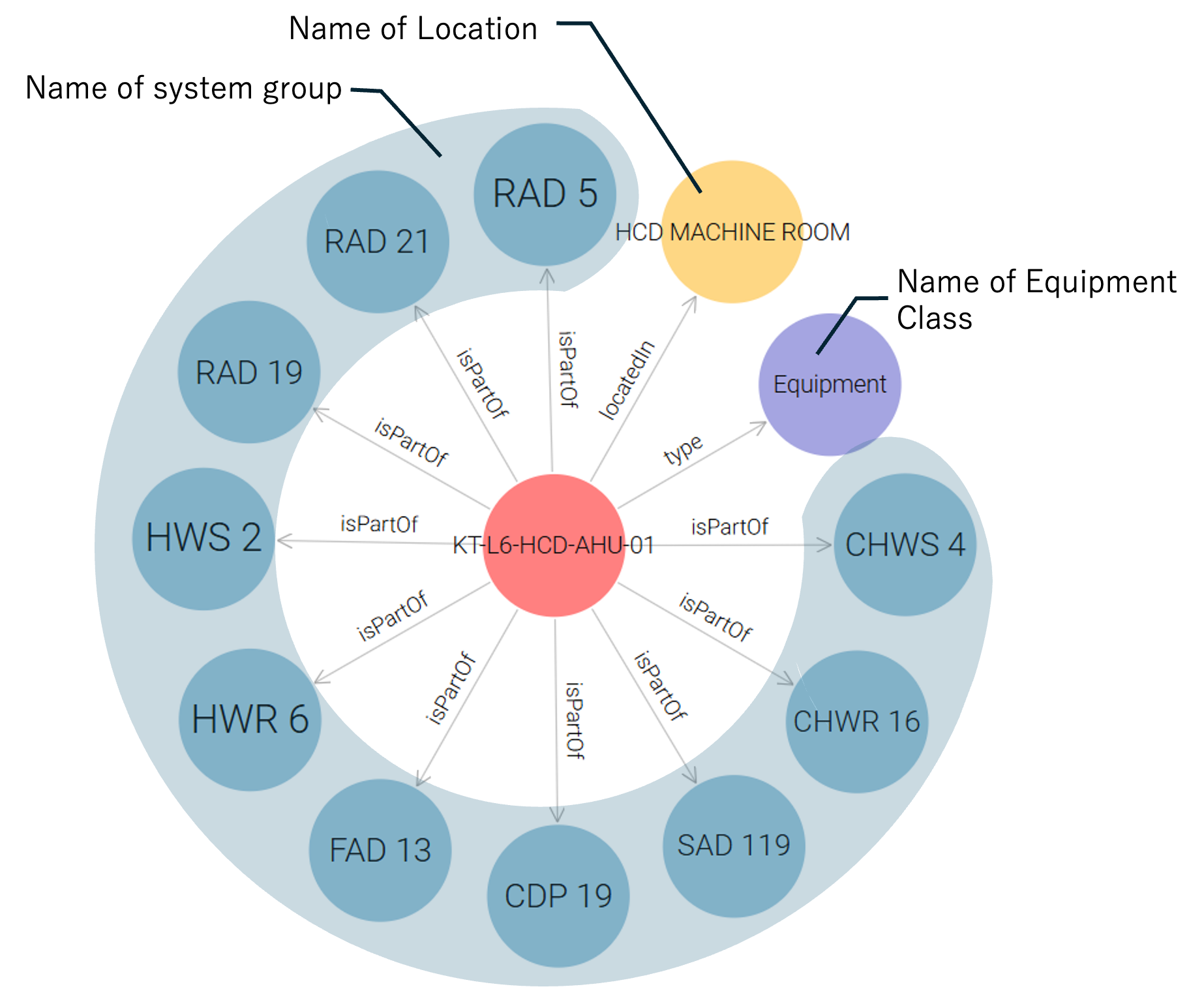}}%
      \hfill
      \raisebox{-0.5\height}{$\xmapsto{\,f\,}$}%
      \hfill
      \raisebox{-0.5\height}{\includegraphics[height=\figh]{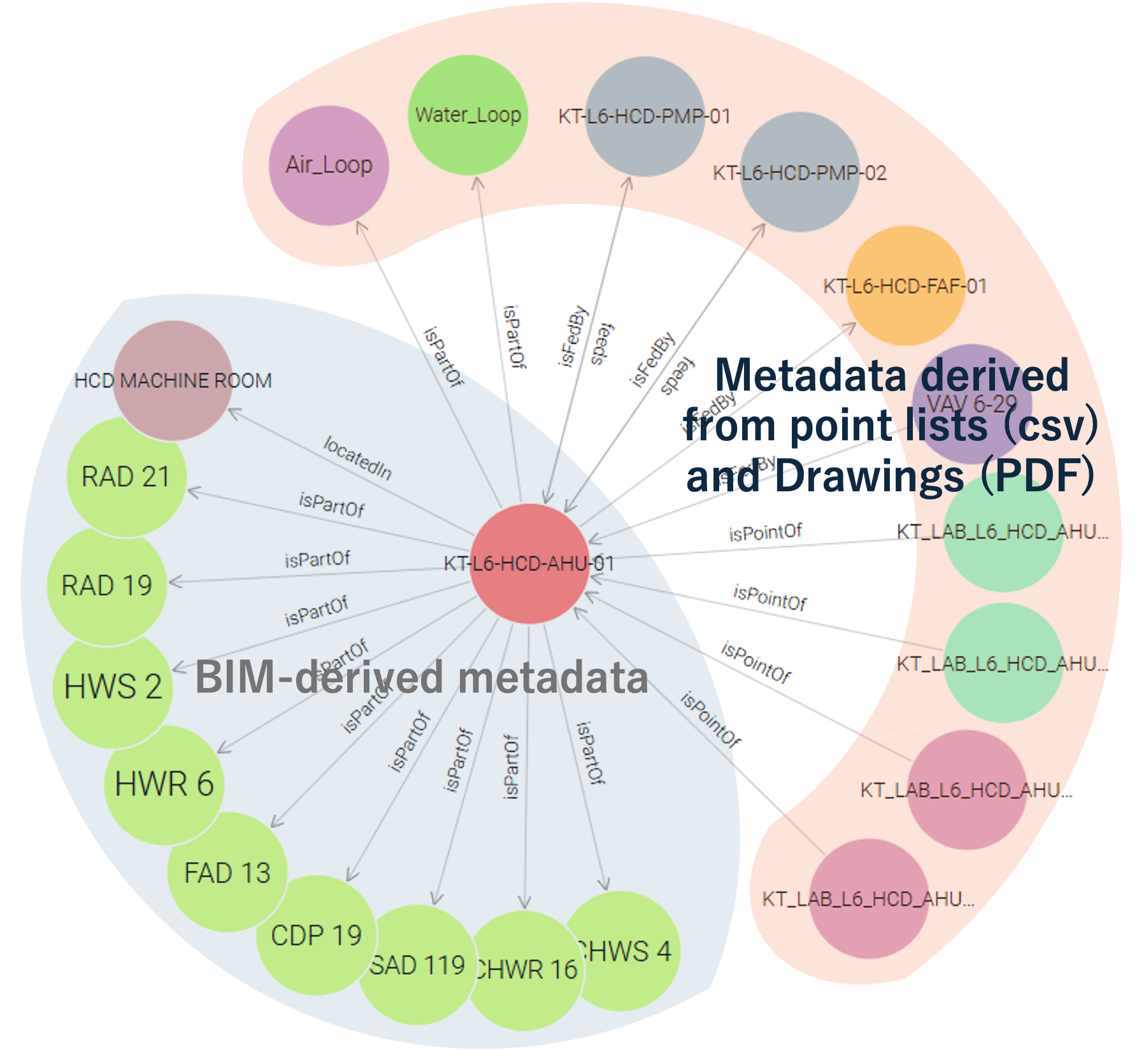}}%
    }
    \caption{The completion task.}
    \label{fig:goal}
  \end{subfigure}
  \caption{Brick represents building entities and their relationships as a typed graph
  (\subref{fig:brick}); our task is to recover missing target edges (\subref{fig:goal}).}
  \label{fig:overview}
\end{figure}

The main obstacle in mass adoption is the difficulty in producing Brick models. The typical workflow begins with a graph generated from Building Information Modeling (BIM) \citep{Sacks_Eastman_Lee_Teicholz_2018}, then augments it with a BMS point list. From there, the metadata in hand is incomplete, and an engineer must manually consult mechanical drawings to recover what is missing. Figure~\ref{fig:overview} illustrates this setting, where the BIM and point-list-derived metadata form the observed graph, and the topology an engineer recovers from mechanical drawings is what remains missing.

The edges which remain to be filled after BIM and BMS extraction are not arbitrary. Entity types and spatial containment are given, as BIM exports carry geometry and hierarchy and point labels follow building naming conventions. What remains to be added is topology. The relations \code{feeds}, \code{hasPart}, and \code{hasPoint} describe how equipment connects, what it contains, and which points report on it. Recovering these relations involves tracing dozens of pages of mechanical drawings, which scales poorly to new buildings. This process requires domain-expertise, and must be repeated for every new Brick model. These three relations account for the vast majority of the schematic-reading effort in building a Brick model by hand, and they are the relations we target.

Our key observation is that the partial graph may contain enough structure for edge inference. Although the BIM and point-list-derived graph is incomplete, it carries substantial information about the building, such as entity types, identifiers, spatial structure, and a subset of the observed relationships. We hypothesize that this partial structure is sufficient to recover the missing relations without examining the mechanical drawings, replacing the schematic-reading step entirely. This approach is a well established problem in graph theory and machine learning. A partially constructed Brick model is a multi-relational knowledge graph with missing edges, and filling in missing edges is knowledge graph completion (KGC), a task with a mature body of methods developed on general-purpose benchmarks. These methods have yet to be applied to the buildings domain. We formulate the final integration step accordingly. Given a candidate entity pair $(h, t)$, a model must determine whether the pair is linked and, if so, predict which relation $r$ holds for $  r\in\{\code{feeds},\code{hasPart},\code{hasPoint}\}.$

Our contributions are:

\begin{enumerate}
  \item We present, to the best of our knowledge, the first systematic study of
  knowledge-graph completion for the Brick schema, targeting the relations that dominate manual modeling effort.

  \item We adapt and evaluate three representative approaches under a shared experimental framework: gradient-boosted trees over pairwise features (GBT), Pairwise Brick Relation Ranking (PBRR), and CompGCN.
  \item We evaluate cross-dataset generalization across MORTAR, Kaggle, and Building Time Series data. We find that no method generalizes uniformly: performance depends strongly on the target relation and the dataset, which exposes the limits of aggregate metrics and motivates relation-specific calibration and evaluation.
\end{enumerate}

\section{Related Works}
\subsection{Brick Schema}

Prior to Brick, building metadata was scattered across formats that each addressed different issues. Project Haystack \citep{projecthaystack} gave practitioners a shared tagging vocabulary for points and equipment. IFC/BIM exports represent relational and spatial structure from design and construction. Domain ontologies like DogOnt \citep{bonino2008dogont}
modeled smart-home devices and their interactions more formally, and SAREF-based extensions
for buildings \citep{PovedaVillaln2018ExtendingTS} pushed toward interoperability with the broader Semantic
Web. Brick \citep{balaji2016brick} aimed to fold these into one graph. It keeps Haystack's tagging instinct for naming entity types, borrows IFC containment structure, and expresses the metadata as an RDF/OWL graph so it can be queried rather than parsed per building.

Work has since grown Brick outward rather than automating how a model gets built.
Luo et al.\ \citep{luo2022occupant} extend the schema to occupant metadata, such as behavior, demographics,
adaptive actions, without touching the core equipment/point/space relations, which evidences that the relational structure generalizes reasonably well to new domains. On the data side, MORTAR \citep{10.1145/3366375} addressed a different bottleneck. Rather than extending
the schema, it packaged real deployed Brick models with time-series data so researchers could
develop and evaluate applications without needing access to an actual building. We rely on MORTAR
for this reason (Section~\ref{sec:datasets}). To the best of our knowledge, no prior work has addressed recovering and inferring Brick edges, which today is still done manually for every new building.

\subsection{Knowledge-Graph Completion}

KGC is a well-studied problem, and current methods split into three families. Embedding models learn a fixed-length vector per entity and relation and score candidate triples against those vectors. TransE \citep{bordes2013transe} treats a relation as a translation from head to tail, DistMult \citep{yang2015embeddingentitiesrelationslearning} and ComplEx \citep{trouillon2016complex} capture symmetric and antisymmetric relations through bilinear and complex-valued products, and ConvE \citep{dettmers2018convolutional2dknowledgegraph} and RotatE
\citep{sun2019rotateknowledgegraphembedding} push further with convolutional scoring and relation-as-rotation, respectively. These methods are fast and do well on standard benchmarks, but they share a structural weakness for our setting: an entity's embedding is tied to its identity, thus the models fail to generalize to out-of-distribution entities, which
describe every point and piece of equipment in a new building.

GNN-based methods look to address this gap. Rather than memorizing per-entity vectors, they aggregate information from a node's local neighborhood, so a representation can be computed for an entity the model has never seen so long as its neighborhood provides sufficient structural information. R-GCN
\citep{schlichtkrull2017modelingrelationaldatagraph} was the first to extend graph convolution to
multi-relational graphs, assigning each relation a transformation matrix, growing parameter count with the number of relations. CompGCN
\citep{vashishth2020compgcn}, which we build on in Section~3.4, addresses this by composing entity and
relation embeddings through a single shared transformation instead, which keeps the model small without giving up relation-awareness. A separate, older tradition is symbolic rather than learned: the Path Ranking Algorithm \citep{lao2010pra} scores candidates by the relational paths
connecting head and tail, and rule-mining systems like AMIE+ \citep{galarraga2015amieplus} and
AnyBURL \citep{meilicke2019anyburl} extract Horn rules directly from the graph. These trade
raw accuracy for human-readable rules, which is the design PBRR follows with its hand-designed, relation-conditioned features rather than a learned rule set.

None of the methods above, embedding-based, GNN-based, or symbolic, have previously been evaluated
on Brick, which uses a small, fixed relation vocabulary dominated by three structural relations, and has strong naming regularities in entity IDs that none of these methods are built to exploit.

\section{Methodology}

\subsection{Problem formulation}
\label{sec:problem}

Let $\mathcal{E}$ denote the entities in a building graph and
$\mathcal{R}=\{\code{feeds},\code{hasPart},\code{hasPoint}\}$
the target relations. Given a candidate entity pair $(h,t)$, each model
predicts whether a target relation exists between the two entities and,
if so, which relation $r\in\mathcal{R}$ holds. The final prediction therefore belongs to $\mathcal{R}\cup\{\varnothing\}$, where $\varnothing$ denotes that no target relation is predicted.

We chose three learned models to compare different ways of using information from the Brick graph. PBRR is the simplest and most interpretable of the three. It uses a selected set of Brick-specific features based on ontology information, naming patterns, relation compatibility, and local graph context. This allows us to test how well a smaller, deliberately chosen feature representation can predict missing relations. GBT preserves the same pairwise prediction setup, but describes each candidate pair using a much larger set of features. This gives the model more information and allows it to learn more detailed associations between entity types and naming patterns. CompGCN takes a different approach by learning directly from the graph structure. Instead of explicitly describing the neighbourhood around each entity through selected features, it learns structural patterns and entity relationships from the surrounding graph through message passing. Together, the three models form a natural progression from selected Brick-specific features, to richer pairwise feature learning, and finally to learning directly from the graph itself.

\subsection{Pairwise Brick Relation Ranking}

PBRR tests whether a 48-dimensional set of selected Brick features is sufficient for predicting missing relations. Brick defines a structured ontology for representing building entities and their relationships,
including equipment, points, locations, and their connections
\citep{balaji2016brick,fierro2020formalizing}. We therefore represent each candidate relation using Brick type information, identifier similarity, nearby observed relationships, and whether the relation is compatible with the entity types.

Because \code{feeds}, \code{hasPart}, and \code{hasPoint} represent different types of relationships and apply differently across Brick entity types \citep{fierro2020formalizing}, PBRR trains a separate one-vs-rest logistic classifier for each relation. For candidate pair $(h,t)$ and relation $r$,
the classifier computes
\begin{equation}
z_r(h,t)=\mathbf{w}_r^{\top}\mathbf{f}(h,r,t)+b_r,
\qquad
P_r(h,t)=\sigma\!\left(z_r(h,t)\right),
\end{equation}
where $\mathbf{f}(h,r,t)$ is the feature vector for candidate pair $(h,t)$
under relation $r$.

Positive examples are observed target edges, while negatives are sampled
from candidate pairs with no known target edge. Since negative candidates substantially outnumber positive edges, the
classification problem is highly imbalanced. This can reduce the classifier's ability to identify positive examples reliably with standard logistic regression
\citep{king2001logistic}. To account for this imbalance, we weight the positive class during training.
We otherwise use the standard $L_2$-regularized logistic classification
objective \citep{hastie2009elements}, giving
\begin{equation}
\mathcal{L}_r =
-\sum_i \left[
\alpha_r y_{ir}\log P_r(h_i,t_i)
+
(1-y_{ir})\log\left(1-P_r(h_i,t_i)\right)
\right]
+
\frac{\lambda}{2}\lVert\mathbf{w}_r\rVert_2^2,
\end{equation}
where $\alpha_r=N_r^-/N_r^+$ is the positive class weighting term, set
according to the negative to positive training ratio, and $\lambda$ controls
the strength of the $L_2$ regularization. A relation-specific threshold $\tau_r$ is selected on the validation set by
maximizing F1. At inference time, PBRR returns the highest-scoring compatible
relation only if its probability exceeds the corresponding threshold.

\subsection{Gradient-boosted trees}

We use a single \code{HistGradientBoostingClassifier} \citep{friedman2001greedy} to score each
candidate triple $(h,r,t)$. Let $\mathbf{x}_e\in\{0,1\}^{C}$ be the multi-hot Brick type encoding
of entity $e$, including all superclasses in the Brick hierarchy so that unseen leaf classes still
activate trained ancestors. The feature map is
\begin{equation}
\phi(h,r,t)=
\bigl[\,\mathbf{x}_h \;;\; \mathbf{x}_t \;;\; \mathbf{x}_h\odot\mathbf{x}_t \;;\;
\lvert \mathbf{x}_h-\mathbf{x}_t\rvert \;;\; \mathbf{g}(h,t) \;;\; \mathbf{e}_r \,\bigr],
\end{equation}
where $\odot$ is the element-wise product, $\mathbf{g}(h,t)\in\mathbb{R}^{9}$ collects directional
string-similarity statistics between the two identifiers (shared tokens, prefixes, and numeric
suffixes), and $\mathbf{e}_r\in\{0,1\}^{3}$ one-hot encodes the candidate relation. The product
and absolute-difference blocks supply type interactions the trees cannot otherwise form. With
$C=126$ this gives $4C+9+3=516$ dimensions.

One model is trained over all three relations, with $\mathbf{e}_r$ permitting relation-specific
behaviour while sharing statistical strength---unlike PBRR, which fits one classifier per
relation. At inference it is queried once per relation to obtain $s_r(h,t)=f(\phi(h,r,t))$, and
the pair takes the highest-scoring relation clearing its own validation-tuned threshold $\tau_r$:
\begin{equation}
\hat{r}(h,t)=\operatorname*{arg\,max}_{r\in\mathcal{A}(h,t)} s_r(h,t),
\quad
\mathcal{A}(h,t)=\{r\in\mathcal{R} : s_r(h,t)\ge\tau_r\},
\end{equation}
predicting $\varnothing$ when $\mathcal{A}(h,t)=\emptyset$. Because $s_r$ is a bounded
probability estimate rather than a calibrated probability, $\tau_r$ is tuned rather than fixed
at $0.5$.

\subsection{CompGCN}
A graph neural network (GNN) operates directly over the graph structure via message passing, in
which each node iteratively aggregates information from its neighbours. Unlike embedding and
feature-based approaches, a GNN learns structural patterns rather than entity identities, which in
principle allows it to generalize to buildings not seen during training: whether a missing edge
should exist is often inferable from the shape of a node's local neighbourhood alone, even without
lexical cues from entity names.

We selected CompGCN \citep{vashishth2020compgcn} for two reasons. First, it treats relations as
first-class learned embeddings rather than fixed identifiers, enabling richer entity-relation
interaction than relation-agnostic GCN variants. It also uses a single shared weight matrix with a
composition operation rather than one matrix per relation, making it parameter-efficient at
approximately 300{,}000 parameters, modest relative to comparable relational GNNs
\citep{schlichtkrull2017modelingrelationaldatagraph}. 

The core message-passing update is
\[
\mathbf{h}_v = f\!\left(\sum_{(u,r)\,\in\,\mathcal{N}(v)} \mathbf{W}_{\lambda(r)}\,
\phi\!\left(\mathbf{x}_u,\, \mathbf{z}_r\right)\right), \qquad
\mathbf{h}_r = \mathbf{W}_{\text{rel}}\,\mathbf{z}_r,
\]
where $\mathbf{x}_u \in \mathbb{R}^d$ are the initial features for entity $u$ (here the mean of its
Brick class token embeddings), $\mathbf{z}_r \in \mathbb{R}^d$ are the initial features for relation
$r$ (randomly initialised and jointly learned), $\phi : \mathbb{R}^d \times \mathbb{R}^d \to
\mathbb{R}^d$ composes a neighbour embedding with its connecting relation embedding before
aggregation, $f$ is a nonlinearity, and $\mathbf{W}_{\lambda(r)} \in \{\mathbf{W}_O, \mathbf{W}_I,
\mathbf{W}_S\}$ are direction-specific weight matrices for original, inverse, and self-loop edges.

The standard CompGCN formulation targets general-purpose link prediction and required four changes
for the Brick setting. First, we replace the standard entity-prediction objective with a joint loss
that directly optimises relation prediction \citep{chen2021relationpredictionauxiliarytraining}:
\[
\arg\max_{\theta \in \Theta} \sum_{\langle h, r, t \rangle \in \mathcal{G}}
\left[\log P_\theta(h \mid r, t) + \log P_\theta(t \mid h, r) + \lambda \log P_\theta(r \mid h, t)\right],
\]
where the first two terms predict the head and tail given the relation, and the third is an
auxiliary relation-prediction term weighted by $\lambda \in \mathbb{R}_+$,
\[
\log P_\theta(r \mid h, t) = \phi_\theta(h, r, t) - \log \sum_{r' \in \mathcal{R}}
\exp\!\left[\phi_\theta(h, r', t)\right],
\]
a softmax over candidate relations given the head--tail pair. Since relation prediction is our
primary task, this term aligns training with inference.

Second, during training we hide relations only, never entities, in contrast to link-prediction
regimes that mask entities, again to match the downstream task. Third, we add a dedicated binary
head for link existence, conditioned on a name-similarity feature between head and tail so that
lexical signals can inform the existence decision where connectivity is ambiguous; separating the
two heads improved end-to-end performance by letting the existence head attend to lexical cues
while the relation head focuses on structure and type. Fourth, node features are initialised
inductively from Brick class embeddings rather than entity-specific vectors, so the same Brick
class is represented consistently across buildings.

\section{Experimental Setup}
\subsection{Datasets and preprocessing}
\label{sec:datasets}

We train our models on MORTAR \citep{10.1145/3366375}, which contains real commercial-building Brick graphs and supplies training, validation, and held-out in-domain buildings. After preprocessing, the combined graph has approximately 19,388 entities, seven core relation types, and 69,159 triples. We additionally evaluate out of distribution on the Kaggle Brick models and Building Time Series (BTS) datasets \citep{Miller_2021, prabowo2024btsbuildingtimeseriesdataset}. The benchmark contains 755 MORTAR test edges, 15,507 Kaggle test edges, and 534 BTS test edges.

We normalize inverse relations (\code{isPointOf}, \code{isPartOf}, and \code{isFedBy}) to their canonical forward directions and map equivalent predicates to Brick. The normalized graphs contain relations beyond our three targets, which we retain as visible context but do not predict. These are entity class assignments, human-readable labels, external references linking points to their BMS timeseries streams, and literal-valued attributes such as floor area. We exclude them from the prediction task as each is previously supplied. Class assignments and labels come from the BMS point list, timeseries references from the BMS itself, and dimensional attributes from the BIM export. Only \code{feeds}, \code{hasPart}, and \code{hasPoint} require an engineer to consult mechanical drawings. Splits are made at the building level to prevent entity and edge-level leakage. Within an evaluation building, visible context edges remain available to structure-aware models, while held-out target edges define the prediction task.

All experiments were run on a single NVIDIA RTX 4090, with the non-neural models on CPU. Each model is
trained with three seeds while the building split and masking are held fixed.

\begin{table}[t]
  \centering
  \footnotesize
  \setlength{\tabcolsep}{3.5pt}
  \begin{tabular}{lrrrrr!{\vrule width 0.4pt}rrr}
    \toprule
    & \multicolumn{5}{c!{\vrule width 0.4pt}}{Dataset summary}
    & \multicolumn{3}{c}{Target relations} \\
    \cmidrule(l){2-6}
    \cmidrule(r){7-9}
    Dataset & Buildings & Entities & Triples & Rel. types & Test edges
    & \code{feeds} & \code{hasPart} & \code{hasPoint} \\
    \midrule
    MORTAR & 45 & 19{,}388 & 69{,}159 & 7 & 755
    & 4{,}650 (21\%) & 4{,}891 (22\%) & 12{,}646 (57\%) \\
    Kaggle & 6 & 23{,}578 & 86{,}493 & 12 & 15{,}507
    & 4{,}639 (12\%) & 3{,}402 (9\%) & 30{,}734 (79\%) \\
    BTS & 1 & 1{,}115 & 4{,}253 & 9 & 534
    & 189 (14\%) & 253 (19\%) & 894 (67\%) \\
    \bottomrule
  \end{tabular}
  \vspace{2pt}
  \caption{Dataset statistics and target-relation distributions after normalization.
  Relation entries report counts with row-wise percentages in parentheses.}
  \label{tab:dataset-stats}
\end{table}

\subsection{Metrics}
Every model emits a link score and a relation distribution for each candidate pair. A candidate pair is predicted linked if its score exceeds a frozen threshold, in which we take the argmax over relations; otherwise no edge is predicted. Positive edges are held-out target edges and negatives are type-constrained false edges of the datasets, sampled at ten per positive edge. Precision-based metrics therefore depend on that ratio.

We report Link F1 over the binary task of determining if a link exists between two entities, agnostic of which relation is the final prediction. Relation classification is scored separately, with counts defined per relation $r$ over four labels: \code{feeds}, \code{hasPart}, \code{hasPoint}, and $\varnothing$, where $\varnothing$ denotes the
prediction of no edge. A true positive $\mathrm{TP}_r$ is a candidate predicted $r$ whose true
label is $r$. A false positive $\mathrm{FP}_r$ is a candidate predicted $r$ whose true label is
$\varnothing$ or another relation. A false negative $\mathrm{FN}_r$ is a candidate whose true label is $r$ but which is predicted $\varnothing$ or
another relation. A true negative $\mathrm{TN}_r$ is a candidate
predicted $\varnothing$ or another relation whose true label is likewise $\varnothing$ or another
relation. The reported F1 scores follow from these counts in the usual way, with micro-F1 pooling
them across the three target relations and macro-F1 averaging the per-relation scores unweighted.
AUPRC is instead computed directly from the link score and is the only threshold-free metric we report.

\subsection{Baselines}
We evaluate two baseline models that bound the learned models. Both are scored on the same candidate sets as the learned models, and their thresholds are set by the same validation rule and frozen in the same way.
\begin{enumerate}
  \item Random: scores every candidate uniformly at random, giving the value of each metric
  under chance. Since this depends on how many candidates are positive, it is a per-dataset reference rather than a fixed number.
  \item Type-pair: uses Brick class alone. For a candidate pair it returns the relation frequencies observed for that class pair in training; if the pair was never seen, it defers to the frequencies for the corresponding top-level Brick classes, and then to the overall training frequencies. Its link score is the largest of the three relation probabilities.
\end{enumerate}

\section{Results}
\label{sec:results}

\begin{table}[t]
  \centering
  \footnotesize
  \setlength{\tabcolsep}{4pt}
  \begin{tabular}{llrrrr}
    \toprule
    Dataset & Model & Link F1 & Macro-F1 & Micro-F1 & AUPRC \\
    \midrule
    MORTAR & Random & $0.208\!\pm\!0.001$ & $0.068\!\pm\!0.002$ & $0.070\!\pm\!0.002$ & $0.115\!\pm\!0.003$ \\
      & Type-pair & $0.255$ & $0.262$ & $0.255$ & $0.230$ \\
      & GBT & $0.413\!\pm\!0.043$ & $0.485\!\pm\!0.050$ & $0.413\!\pm\!0.043$ & $0.472\!\pm\!0.008$ \\
      & PBRR & $\mathbf{0.824\!\pm\!0.044}$ & $0.739\!\pm\!0.021$ & $\mathbf{0.824\!\pm\!0.044}$ & $0.815\!\pm\!0.018$ \\
      & CompGCN & $0.810\!\pm\!0.031$ & $\mathbf{0.746\!\pm\!0.022}$ & $0.810\!\pm\!0.031$ & $\mathbf{0.916\!\pm\!0.007}$ \\
    \midrule
    Kaggle & Random & $0.183\!\pm\!0.001$ & $0.056\!\pm\!0.001$ & $0.061\!\pm\!0.001$ & $0.102\!\pm\!0.000$ \\
      & Type-pair & $0.158$ & $0.160$ & $0.151$ & $0.104$ \\
      & GBT & $0.171\!\pm\!0.003$ & $0.158\!\pm\!0.005$ & $0.146\!\pm\!0.004$ & $0.130\!\pm\!0.001$ \\
      & PBRR & $0.059\!\pm\!0.083$ & $0.005\!\pm\!0.006$ & $0.008\!\pm\!0.013$ & $0.110\!\pm\!0.002$ \\
      & CompGCN & $\mathbf{0.407\!\pm\!0.033}$ & $\mathbf{0.173\!\pm\!0.020}$ & $\mathbf{0.223\!\pm\!0.036}$ & $\mathbf{0.510\!\pm\!0.031}$ \\
    \midrule
    BTS & Random & $0.207\!\pm\!0.001$ & $0.065\!\pm\!0.002$ & $0.068\!\pm\!0.001$ & $0.114\!\pm\!0.009$ \\
      & Type-pair & $0.185$ & $0.172$ & $0.166$ & $0.112$ \\
      & GBT & $\mathbf{0.608\!\pm\!0.002}$ & $\mathbf{0.250\!\pm\!0.006}$ & $\mathbf{0.561\!\pm\!0.003}$ & $\mathbf{0.545\!\pm\!0.005}$ \\
      & PBRR & $0.092\!\pm\!0.089$ & $0.064\!\pm\!0.035$ & $0.059\!\pm\!0.040$ & $0.142\!\pm\!0.004$ \\
      & CompGCN & $0.228\!\pm\!0.019$ & $0.082\!\pm\!0.007$ & $0.071\!\pm\!0.009$ & $0.512\!\pm\!0.014$ \\
    \bottomrule
  \end{tabular}
  \vspace{3pt}
  \caption{Aggregate link-existence and relation-classification performance across MORTAR, Kaggle, and BTS. Values are mean $\pm$ standard deviation. Best results per dataset are bold.}
  \label{tab:three-dataset-results}
\end{table}

\begin{table}[t]
\centering
\small
\setlength{\tabcolsep}{4pt}
\renewcommand{\arraystretch}{1.05}

\begin{tabular}{lccc|ccc|ccc}
\toprule
& \multicolumn{3}{c|}{MORTAR}
& \multicolumn{3}{c|}{Kaggle}
& \multicolumn{3}{c}{BTS} \\
Model
& hasPoint & hasPart & feeds
& hasPoint & hasPart & feeds
& hasPoint & hasPart & feeds \\
\midrule

Random
& 0.107 & 0.046 & 0.052
& 0.129 & 0.017 & 0.023
& 0.124 & 0.041 & 0.030 \\

Type-pair
& 0.257 & 0.310 & 0.219
& 0.148 & \textbf{0.178} & 0.154
& 0.173 & \textbf{0.234} & \textbf{0.109} \\

GBT
& 0.282 & 0.652 & 0.521
& 0.131 & 0.158 & \textbf{0.186}
& \textbf{0.743} & 0.000 & 0.009 \\

PBRR
& 0.927 & 0.658 & \textbf{0.633}
& 0.000 & 0.013 & 0.002
& 0.000 & 0.126 & 0.067 \\

CompGCN
& \textbf{0.944} & \textbf{0.668} & 0.624
& \textbf{0.342} & 0.062 & 0.116
& 0.159 & 0.070 & 0.016 \\

\bottomrule
\end{tabular}
\vspace{2pt}

\caption{End-to-end F1 for each target relation across datasets. Values are means across seeds; best results per dataset and relation are bold.}
\label{tab:end-to-end}

\end{table}

Table~\ref{tab:three-dataset-results} reports results by dataset and Table~\ref{tab:end-to-end} breaks them down by target relation. On the in-distribution MORTAR test set, PBRR and CompGCN both perform well. PBRR has the highest Link F1 and micro-F1, CompGCN the highest macro-F1, and both gaps are smaller than the seed-to-seed standard deviation. The clearer separation is in AUPRC, the only threshold-free metric we report (valuable given the ten to one ratio of negative to positive edges), where CompGCN reaches $0.916$ against PBRR's $0.815$. MORTAR is also the only dataset on which all models perform well across all three relations rather than relying on one. This establishes the feasibility of inferring missing relations from information present in a partially constructed graph, with no mechanical drawings supplied to any model.

Each model degrades out of distribution. CompGCN leads every metric on
Kaggle, but its macro-F1 of $0.173$ barely exceeds the type-pair baseline's
(Table~\ref{tab:three-dataset-results}). CompGCN's Kaggle performance is
almost entirely driven by \code{hasPoint} ($0.342$;
Table~\ref{tab:end-to-end}). GBT shows a similar pattern on BTS, where it
leads every aggregate metric on the strength of a \code{hasPoint} F1 of
$0.743$, while scoring $0.000$ on \code{hasPart} and $0.009$ on
\code{feeds} (Table~\ref{tab:end-to-end}). In both cases, the aggregate
performance is dominated by a single relation rather than balanced
completion. PBRR also fails to generalize to out of distribution datasets, with
Link F1 falling from $0.824$ on MORTAR to $0.059$ on Kaggle and $0.092$ on
BTS (Table~\ref{tab:three-dataset-results}). This sharp decline suggests that
the information PBRR relies on is highly relevant to the way relations are
represented in MORTAR, but does not transfer consistently when those
relationships are encoded differently across Kaggle and BTS. GBT's unusually strong BTS result is
likewise driven by dataset composition rather than broader generalisation.
Its advantage is confined to \code{hasPoint} (F1 $0.743$ versus $0.000$ and
$0.009$ on the other two relations; Table~\ref{tab:end-to-end}), which
accounts for $67\%$ of BTS target edges
(Table~\ref{tab:dataset-stats}). This relation is also lexically
easier in BTS than in MORTAR because BTS point URIs are namespaced under
their parent equipment identifiers, giving GBT's string-overlap features a
stronger out-of-distribution signal.

Two more observations complicate declaring a winning model. First, CompGCN's AUPRC stays high on both out-of-distribution datasets while its Link F1 falls to $0.407$ and $0.228$, so its candidate ordering transfers considerably better than its frozen threshold. Second, the type-pair baseline performs best in three of the six out-of-distribution dataset-relation combinations, meaning Brick class alone outperforms every learned model on \code{hasPart} for both unseen datasets and on \code{feeds} for BTS.

\section{Conclusion}

We apply knowledge graph completion methods to the problem of completing Brick schema graphs. To the best of our knowledge, this is the first application of such methods to this domain. We compare GBT, PBRR, and CompGCN for predicting relations in incomplete Brick graphs. On MORTAR, both PBRR and CompGCN recover missing relations at F1 well above baselines across all three target relations, using only the entity types, identifiers, and partial structure present after BIM and BMS extraction. We conclude that the information needed to complete a Brick model is substantially present in the partial model.

On Kaggle, CompGCN ranks edges most effectively but its relation performance is driven by \code{hasPoint}. On BTS, GBT's aggregate lead is also almost solely a \code{hasPoint} result, while the type-pair baseline is best for \code{hasPart} and \code{feeds}. No method generalizes across all three datasets, and the model that wins on aggregate metrics rarely wins on all relations. We conclude that Brick completion is not well served by a single global architecture. Recoverability depends not only on the model architecture but also on which signals (types, identifiers, or graph structure) are consistently encoded in the target dataset; a practical system should characterize the target graph first and select the model that suits it. We hope this benchmark encourages further work on automating Brick model construction.

\section*{Acknowledgments}
We thank the Institute for Pure and Applied Mathematics (IPAM) Research in Industrial Projects for Students (RIPS) program and the Institute for Mathematical Sciences at the National University of Singapore for hosting this research, and Kajima Corporation for sponsoring this project.

\renewcommand{\bibfont}{\small}
\bibliographystyle{plainnat}
\bibliography{references}

@inproceedings{balaji2016brick,
  author    = {Bharathan Balaji and Arka Bhattacharya and Gabriel Fierro and Jingkun Gao and Joshua Gluck and Dezhi Hong and Aslak Johansen and Jason Koh and Joern Ploennigs and Yuvraj Agarwal and Mario Berg{\'e}s and David Culler and Rajesh Gupta and Mikkel Baun Kj{\ae}rgaard and Mani Srivastava and Kamin Whitehouse},
  title     = {Brick: Towards a Unified Metadata Schema for Buildings},
  booktitle = {Proceedings of the 3rd ACM International Conference on Systems for Energy-Efficient Built Environments},
  pages     = {41--50},
  year      = {2016},
  doi       = {10.1145/2993422.2993577}
}

@article{10.1145/3366375,
author = {Fierro, Gabe and Pritoni, Marco and Abdelbaky, Moustafa and Lengyel, Daniel and Leyden, John and Prakash, Anand and Gupta, Pranav and Raftery, Paul and Peffer, Therese and Thomson, Greg and Culler, David E.},
title = {Mortar: An Open Testbed for Portable Building Analytics},
year = {2019},
issue_date = {February 2020},
publisher = {Association for Computing Machinery},
address = {New York, NY, USA},
volume = {16},
number = {1},
issn = {1550-4859},
url = {https://doi.org/10.1145/3366375},
doi = {10.1145/3366375},
journal = {ACM Trans. Sen. Netw.},
month = dec,
articleno = {7},
numpages = {31}
}

@article{friedman2001greedy,
  author  = {Jerome H. Friedman},
  title   = {Greedy Function Approximation: A Gradient Boosting Machine},
  journal = {The Annals of Statistics},
  volume  = {29},
  number  = {5},
  pages   = {1189--1232},
  year    = {2001},
  doi     = {10.1214/aos/1013203451}
}

@article{galarraga2015amieplus,
  author  = {Luis Gal{\'a}rraga and Christina Teflioudi and Katja Hose and Fabian M. Suchanek},
  title   = {Fast Rule Mining in Ontological Knowledge Bases with {AMIE+}},
  journal = {The VLDB Journal},
  volume  = {24},
  number  = {6},
  year    = {2015},
  doi     = {10.1007/s00778-015-0394-1}
}

@inproceedings{meilicke2019anyburl,
  author    = {Christian Meilicke and Melisachew Wudage Chekol and Daniel Ruffinelli and Heiner Stuckenschmidt},
  title     = {Anytime Bottom-Up Rule Learning for Knowledge Graph Completion},
  booktitle = {Proceedings of the Twenty-Eighth International Joint Conference on Artificial Intelligence},
  pages     = {3137--3143},
  year      = {2019},
  doi       = {10.24963/ijcai.2019/435}
}

@misc{vashishth2020compgcn,
  author    = {Shikhar Vashishth and Soumya Sanyal and Vikram Nitin and Partha Talukdar},
  title     = {Composition-based Multi-Relational Graph Convolutional Networks},
  eprint={1911.03082},
  archivePrefix={arXiv},
  primaryClass={cs.LG},
  year      = {2020},
  url={https://arxiv.org/abs/1911.03082}
}

@inproceedings{bordes2013transe,
  author  = {Antoine Bordes and Nicolas Usunier and Alberto Garcia-Duran and Jason Weston and Oksana Yakhnenko},
  title   = {Translating Embeddings for Modeling Multi-relational Data},
  booktitle = {Proceedings of the 27th International Conference on Neural Information Processing Systems - Volume 2},
  pages = {2787–2795},
  numpages = {9},
  location = {Lake Tahoe, Nevada},
  series = {NIPS'13},
  year    = {2013}
}

@misc{yang2015embeddingentitiesrelationslearning,
  author    = {Bishan Yang and Wen-tau Yih and Xiaodong He and Jianfeng Gao and Li Deng},
  title     = {Embedding Entities and Relations for Learning and Inference in Knowledge Bases},
  eprint={1412.6575},
  archivePrefix={arXiv},
  primaryClass={cs.CL},
  url={https://arxiv.org/abs/1412.6575}, 
  year      = {2015}
}

@misc{trouillon2016complex,
  author    = {Th{\'e}o Trouillon and Johannes Welbl and Sebastian Riedel and {\'E}ric Gaussier and Guillaume Bouchard},
  title     = {Complex Embeddings for Simple Link Prediction},
  eprint={1606.06357},
  archivePrefix={arXiv},
  primaryClass={cs.AI},
  url={https://arxiv.org/abs/1606.06357}, 
  year      = {2016}
}

@misc{dettmers2018convolutional2dknowledgegraph,
  author    = {Tim Dettmers and Pasquale Minervini and Pontus Stenetorp and Sebastian Riedel},
  title     = {Convolutional 2D Knowledge Graph Embeddings},
  eprint={1707.01476},
  archivePrefix={arXiv},
  primaryClass={cs.LG},
  url={https://arxiv.org/abs/1707.01476},
  year      = {2018}
}

@misc{sun2019rotateknowledgegraphembedding,
      title={RotatE: Knowledge Graph Embedding by Relational Rotation in Complex Space}, 
      author={Zhiqing Sun and Zhi-Hong Deng and Jian-Yun Nie and Jian Tang},
      year={2019},
      eprint={1902.10197},
      archivePrefix={arXiv},
      primaryClass={cs.LG},
      url={https://arxiv.org/abs/1902.10197}, 
}

@misc{schlichtkrull2017modelingrelationaldatagraph,
      title={Modeling Relational Data with Graph Convolutional Networks}, 
      author={Michael Schlichtkrull and Thomas N. Kipf and Peter Bloem and Rianne van den Berg and Ivan Titov and Max Welling},
      year={2017},
      eprint={1703.06103},
      archivePrefix={arXiv},
      primaryClass={stat.ML},
      url={https://arxiv.org/abs/1703.06103}, 
}

@article{lao2010pra,
  author  = {Ni Lao and William W. Cohen},
  title   = {Relational Retrieval Using a Combination of Path-Constrained Random Walks},
  journal = {Machine Learning},
  volume  = {81},
  number  = {1},
  pages   = {53--67},
  year    = {2010}
}

@misc{chen2021relationpredictionauxiliarytraining,
      title={Relation Prediction as an Auxiliary Training Objective for Improving Multi-Relational Graph Representations}, 
      author={Yihong Chen and Pasquale Minervini and Sebastian Riedel and Pontus Stenetorp},
      year={2021},
      eprint={2110.02834},
      archivePrefix={arXiv},
      primaryClass={cs.CL},
      url={https://arxiv.org/abs/2110.02834}, 
}

@inproceedings{bonino2008dogont,
  author    = {Dario Bonino and Fulvio Corno},
  title     = {DogOnt - Ontology Modeling for Intelligent Domotic Environments},
  booktitle = {Proceedings of the 7th International Conference on The Semantic Web},
  pages     = {790--803},
  year      = {2008},
  url = {https://doi.org/10.1007/978-3-540-88564-1_51},
  doi = {10.1007/978-3-540-88564-1_51},
}

@inproceedings{PovedaVillaln2018ExtendingTS,
  title={Extending the SAREF ontology for building devices and topology},
  author={Mar{\'i}a Poveda-Villal{\'o}n and Ra{\'u}l Garc{\'i}a-Castro},
  booktitle={LDAC},
  year={2018},
  url={https://api.semanticscholar.org/CorpusID:53551935}
}

@article{luo2022occupant,
  author  = {Na Luo and Gabe Fierro and Yapan Liu and Bing Dong and Tianzhen Hong},
  title   = {Extending the Brick Schema to Represent Metadata of Occupants},
  journal = {Automation in Construction},
  volume  = {139},
  pages   = {104307},
  year    = {2022},
  doi = {https://doi.org/10.1016/j.autcon.2022.104307},
  url = {https://www.sciencedirect.com/science/article/pii/S0926580522001807}
}

@article{pritoni2021metadata,
  author  = {Pritoni, Marco and Paine, Drew and Fierro, Gabe and Mosiman, Cory
             and Poplawski, Michael and Saha, Avijit and Bender, Joel},
  title   = {Metadata Schemas and Ontologies for Building Energy Applications:
             A Critical Review and Use Case Analysis},
  journal = {Energies},
  volume  = {14}, number = {7}, ARTICLE-NUMBER = {2024}, year = {2021},
  doi     = {10.3390/en14072024},
  URL = {https://www.mdpi.com/1996-1073/14/7/2024}
}

@article{fierro2020formalizing,
author = {Fierro, Gabe and Koh, Jason and Nagare, Shreyas and Zang, Xiaolin
          and Agarwal, Yuvraj and Gupta, Rajesh K. and Culler, David E.},
title = {Formalizing Tag-Based Metadata With the Brick Ontology},
journal = {Frontiers in Built Environment},
volume = {Volume 6 - 2020},
year = {2020},
doi = {10.3389/fbuil.2020.558034},
url = {https://www.frontiersin.org/journals/built-environment/articles/10.3389/fbuil.2020.558034},
issn = {2297-3362}
}

@article{king2001logistic,
  title   = {Logistic Regression in Rare Events Data},
  author  = {King, Gary and Zeng, Langche},
  journal = {Political Analysis},
  volume  = {9},
  number  = {2},
  pages   = {137--163},
  year    = {2001},
  doi     = {10.1093/oxfordjournals.pan.a004868}
}

@book{hastie2009elements,
  author    = {Hastie, Trevor and Tibshirani, Robert and Friedman, Jerome},
  title     = {The Elements of Statistical Learning: Data Mining, Inference, and Prediction},
  edition   = {2},
  series    = {Springer Series in Statistics},
  publisher = {Springer},
  address   = {New York, NY},
  year      = {2009},
  doi       = {10.1007/978-0-387-84858-7},
  isbn      = {978-0-387-84858-7}
}

@misc{projecthaystack,
  title        = {Project Haystack},
  url          = {https://project-haystack.org/},
}

@misc{prabowo2024btsbuildingtimeseriesdataset,
      title={BTS: Building Timeseries Dataset: Empowering Large-Scale Building Analytics}, 
      author={Arian Prabowo and Xiachong Lin and Imran Razzak and Hao Xue and Emily W. Yap and Matthew Amos and Flora D. Salim},
      year={2024},
      eprint={2406.08990},
      archivePrefix={arXiv},
      primaryClass={cs.LG},
      url={https://arxiv.org/abs/2406.08990}, 
}

@misc{Miller_2021, 
    title={Reference building BMS data brick models}, 
    url={https://www.kaggle.com/datasets/claytonmiller/reference-building-bms-data-brick-models}, 
    journal={Kaggle}, 
    author={Miller, Clayton}, 
    year={2021}, 
    month={Sep}
}

@book{Sacks_Eastman_Lee_Teicholz_2018, place={Hoboken, NJ}, title = {BIM Handbook: A Guide to Building Information Modeling for Owners, Managers, Designers, Engineers and Contractors}, publisher={John Wiley \& Sons, Inc}, author = {Eastman, Chuck and Teicholz, Paul and Sacks, Rafael and Liston, Kathleen}, isbn = {0470185287},
publisher = {Wiley Publishing}, year = {2008}}

\end{document}